\documentclass[conference]{IEEEtran}
\IEEEoverridecommandlockouts

\usepackage{amssymb}
\usepackage{amsmath,amsfonts}
\usepackage{array}
\usepackage[caption=false,font=normalsize,labelfont=sf,textfont=sf]{subfig}
\usepackage{textcomp}
\usepackage{stfloats}
\usepackage{url}
\usepackage{verbatim}
\usepackage{graphicx}
\usepackage{cite}
\usepackage{amsmath}
\usepackage{autobreak}
\usepackage{algorithmic}
\usepackage{booktabs}
\usepackage{bm}
\usepackage{float}
\usepackage{soul}
\usepackage{xcolor}
\usepackage{multirow}
\usepackage{graphicx}

\usepackage[linesnumbered,boxed,ruled,commentsnumbered]{algorithm2e}
\usepackage{textcomp}
\usepackage{threeparttable}

\def\BibTeX{{\rm B\kern-.05em{\sc i\kern-.025em b}\kern-.08em
    T\kern-.1667em\lower.7ex\hbox{E}\kern-.125emX}}
\begin{document}

\title{Retrieval Augmented Generation-Enhanced Distributed LLM Agents for Generalizable Traffic Signal Control with Emergency Vehicles
\thanks{Corresponding author: Lei Li (leili@bupt.edu.cn).}
}

\author{
    \IEEEauthorblockN{
        Xinhang Li\IEEEauthorrefmark{1},
         Qing Guo\IEEEauthorrefmark{1},
        Junyu Chen\IEEEauthorrefmark{1},
        Zheng Guo\IEEEauthorrefmark{1},
        Shengzhe Xu\IEEEauthorrefmark{1},
        Lei Li\IEEEauthorrefmark{1},
        Lin Zhang\IEEEauthorrefmark{1}\IEEEauthorrefmark{2}
    }
    \IEEEauthorblockA{\IEEEauthorrefmark{1} School of  Artificial Intelligence, Beijing University of Posts and Telecommunications, Beijing, China}
    \IEEEauthorblockA{\IEEEauthorrefmark{2}Beijing Big Data Center, Beijing, China}
}

\maketitle

\begin{abstract}
With increasing urban traffic complexity, Traffic Signal Control (TSC) is essential for optimizing traffic flow and improving road safety. Large Language Models (LLMs) emerge as promising approaches for TSC. However, they are prone to hallucinations in emergencies, leading to unreliable decisions that may cause substantial delays for emergency vehicles.
Moreover, diverse intersection types present substantial challenges for traffic state encoding and cross-intersection training, limiting generalization across heterogeneous intersections.
Therefore, this paper proposes Retrieval Augmented Generation (RAG)-enhanced distributed LLM agents with Emergency response for Generalizable TSC (REG-TSC). 
Firstly, this paper presents an emergency-aware reasoning framework, which dynamically adjusts reasoning depth based on the emergency scenario and is equipped with a novel Reviewer-based Emergency RAG (RERAG) to distill specific knowledge and guidance from historical cases, enhancing the reliability and rationality of agents' emergency decisions.
Secondly, this paper designs a type-agnostic traffic representation and proposes a Reward-guided Reinforced Refinement ($\text{R}^3$) for heterogeneous intersections. $\text{R}^3$ adaptively samples training experience from diverse intersections with environment feedback-based priority and fine-tunes LLM agents with a designed reward-weighted likelihood loss, guiding REG-TSC toward high-reward policies across heterogeneous intersections. 
On three real-world road networks with 17 to 177 heterogeneous intersections, extensive experiments show that REG-TSC reduces travel time by 42.00\%, queue length by 62.31\%, and emergency vehicle waiting time by 83.16\%, outperforming other state-of-the-art methods.
\end{abstract}


\section{Introduction}

Traffic Signal Control (TSC) plays a critical role in urban traffic management by improving the traffic efficiency and relieving congestion \cite{a1}. 
Traditional methods, such as transportation-based and Reinforcement Learning (RL)-based methods \cite{a3} have shown great potential in TSC for structured intersections with simple traffic patterns.
The emergence of Large Language Models (LLMs) has partially mitigated the generalization limitations of traditional methods. However, such generalization is mostly limited to variable traffic flow and diverse road network structures with simplified same intersections \cite{a2,a5}. LLMs still struggle to manage heterogeneous intersections and to reason reliably in emergency scenarios. These limitations remain key barriers to the research of LLM-based traffic signal optimization agents. 

Emergency scenarios, due to suddenness and urgency, introduce great complexities to TSC and have become a research focus. \cite{EV1} employed Deep Q-Network (DQN) to balance the impact of emergency vehicles on regular traffic flow. \cite{EV2} and \cite{EV3} both proposed novel vehicle-road coordination schemes to simultaneously manage signal phases and the routing of emergency vehicles. Wang et al. adopted multiple LLM agents to assess traffic phases, prioritize emergency vehicles, and verify rule compliance \cite{EV7}. However, the above methods are tailored specifically for emergency vehicles and are not applicable to generalizable traffic signal optimization. In contrast, several studies consider the occurrence of emergency vehicles as stochastic events in the traffic \cite{EV4,EV5,EV6}, aligning with real-world conditions. In particular, \cite{EV5} and \cite{EV6} leveraged LLMs to assist or refine RL decision-making in emergency scenarios. However, the LLM agents in the above methods are prone to hallucinations and follow fixed reasoning modes across all scenarios, limiting policy reliability and efficiency.

\begin{figure*}[t]
    \centering
    \makebox[\textwidth][c]{ 
        \includegraphics[width=0.92\textwidth]{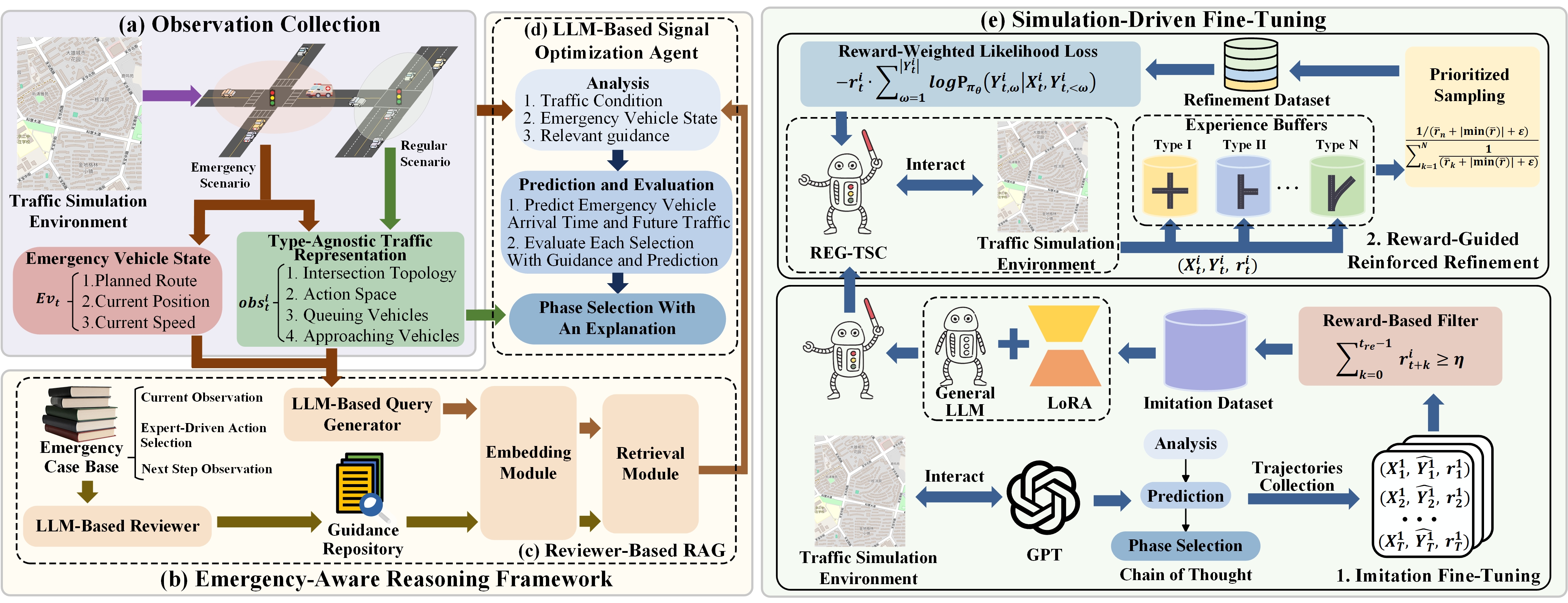} 
    }
    \caption{Architecture of REG-TSC. (a) Observation Collection obtains traffic states and converts them into natural language representations. In (b) Emergency-Aware Reasoning Framework, (c) Reviewer-Based RAG retrieves critical guidance based on the current traffic and emergency vehicle states. (d) LLM-Based Signal Optimization Agent performs deep reasoning by integrating the guidance with traffic representations. (e) Simulation-Driven Fine-Tuning is conducted in two stages: imitation fine-tuning and reward-guided reinforced refinement.}
    \label{fig1}
\end{figure*}

Retrieval Augmented Generation (RAG), as an advanced technique in LLMs, enables the incorporation of domain-specific knowledge to enhance LLMs' performance on specialized tasks \cite{RAG2}. Additionally, with the support of RAG, the hallucination problem of LLMs is mitigated through referenced knowledge \cite{RAG9,RAG6,a6}. For TSC, hallucinations may result in unsafe or inefficient decisions, posing considerable risks \cite{RAG4}. Particularly in safety-critical emergency scenarios, the rationality and reliability of policies are paramount. 
However, existing RAG merely retrieve raw information without effective verification or distillation, unsuitable for highly dynamic urban traffic with emergency vehicles.

In addition, RAG supports the extension of Chain of Thought (CoT), allowing LLMs to perform deep and fine-grained reasoning by extracting additional valuable information from prompts and retrieved knowledge \cite{RAG10}. However, the reasoning with fixed reasoning depth across all intersections is inefficient. Simple scenarios do not require intensive computation, while emergency scenarios demand more nuanced analysis. This one-size-fits-all mode not only wastes computational resources but also hinders real-time decision-making. Therefore, a key challenge lies in balancing computational efficiency with decision-making effectiveness.

Recent RL-based TSC methods have increasingly explored generalization across heterogeneous intersections. 
These studies mainly improve uniform state and action representations \cite{HI2,HI3} and optimize transfer training \cite{HI4,HI6}, aiming to adapt RL to varying intersection layouts. 
Specifically, \cite{HI2} employed attention mechanisms to flexibly extract traffic dynamics at heterogeneous intersections. \cite{HI3} designed an encoder-decoder structure to project intersection states into a unified space. \cite{HI4} proposed a general scenario-agnostic RL framework to conduct a large-scale co-training with multiple scenarios. \cite{HI6} allowed heterogeneous intersections to share Control Knowledge.
However, as an emerging paradigm,  LLMs still face considerable challenges in unified representation and transfer learning across heterogeneous intersections.

In summary, emergency scenarios and heterogeneous intersections remain a major obstacle to complex TSC for LLM-based methods. Therefore, this paper proposes RAG-enhanced distributed LLM agents with Emergency response for Generalizable TSC (REG-TSC), as shown in Fig. \ref{fig1}. The proposed emergency-aware reasoning framework employs a novel Reviewer-based Emergency RAG (RERAG) to efficiently extract knowledge from historical scenarios to form a guidance repository. And the appropriate guidance retrieved from the repository is embedded into the prompts, enabling the LLMs to perform deep reasoning for effective emergency response. Besides, during the training process, we design a Reward-guided Reinforced Refinement ($\text{R}^3$) that dynamically rebalances the training data according to the simulation results and fine-tunes REG-TSC with reward-weighted loss, improving generalization across heterogeneous intersections.
The contributions of this paper are summarized as follows. 
\begin{itemize}
\item This paper proposes REG-TSC for generalizable signal optimization in urban dynamic traffic. Empowered by the emergency-aware reasoning framework, REG-TSC enables LLM agents to perform deep reasoning with RERAG, generating rational and reliable emergency responses with critical knowledge and guidance.

\item This paper designs a type-agnostic traffic representation to standardize static and dynamic features across heterogeneous intersections.
We further propose $\text{R}^3$, which adaptively samples the training dataset with priority based on environmental feedback and refines REG-TSC with a designed reward-weighted likelihood loss, thereby boosting performance across diverse intersection types.

\item This paper constructs three real-world road networks with the number of intersections varying from 17 to 177. Extensive experiments show that REG-TSC reduces travel time by 42.00\%, queue length by 62.31\%, and emergency vehicle waiting time by 83.16\% on average compared with other state-of-the-art (SOTA) methods. 
\end{itemize} 


\section{Generalizable TSC with Emergency Vehicles} \label{sectionII}

\subsection{Road Networks with Heterogeneous Intersections}

Real-world urban road networks consist of various heterogeneous intersections differing in shapes and lane layouts. This paper constructs high-authenticity urban road networks, covering four common intersection shapes with different lane layouts: cross intersections, T-intersections, Y-intersections, and roundabouts. There are $N$ types of heterogeneous intersections. The following are key definitions for representing traffic networks with heterogeneous intersections.

\begin{figure}[t]
\centering
\includegraphics[width=\columnwidth]{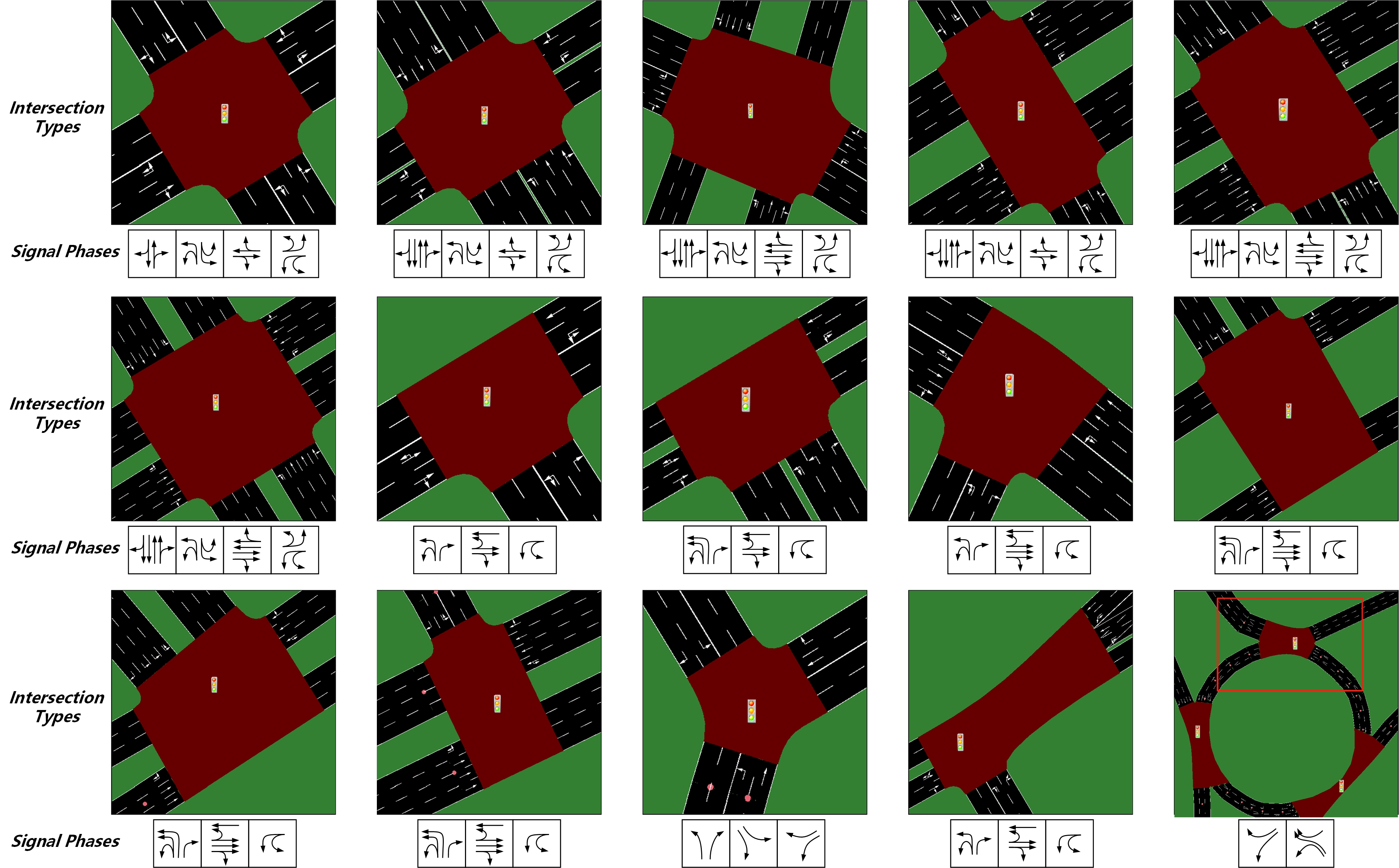}
\caption{An illustration of heterogeneous intersections and signal phases.}
\label{he_inter}
\end{figure}

\textit{Definition 1 (Road Network):} The road network is modeled as a directed graph, where nodes correspond to \textit{I} intersections and edges represent roads. Each road contains a varying number of lanes, and each lane has a fixed direction of travel.

\textit{Definition 2 (Heterogeneous Intersections):} At intersection $i$, there are $UL_i$ upstream lanes and $DL_i$ downstream lanes. Specifically, the upstream lanes through which vehicles enter intersection $i$ are denoted as $\mathcal{UL}_i = \{ ul_i^j \mid j = 1, 2,..., UL_i \}$, and the downstream lanes through which vehicles exit intersection $i$ are denoted as $\mathcal{DL}_i = \{ dl_i^k \mid k = 1, 2,..., DL_i \}$.

\textit{Definition 3 (Traffic Movement):} A traffic movement is defined as a vehicle crossing an intersection from one upstream lane to one downstream lane. A movement from lane $ul_i^j$ to lane $dl_i^k$ is denoted as $(ul_i^j, dl_i^k)$. At intersection $i$, there are $TM_i$ traffic movements $\mathcal{TM}_i \subseteq \mathcal{UL}_i \times \mathcal{DL}_i$.

\textit{Definition 4 (Traffic Signal Phase):} A traffic signal phase is defined as a set of non-conflicting traffic movements that are allowed to travel simultaneously. The set of all phases at intersection $i$ is denoted as $\mathcal{PH}_i = \{ PH_i^m \mid m = 1, 2, \ldots, J_i \}$, where $J_i$ is the total number of signal phases of intersection $i$. 
Fig. \ref{he_inter} illustrates signal phases for different types of heterogeneous intersections.

\subsection{TSC with Emergency Scenarios}

In TSC with sudden emergency scenarios, we consider an urban traffic network consisting of $I$ intersections, where regular traffic flows coexist with emergency vehicles that appear unpredictably. 
The simulation runs for $T$ continuous time steps, during which $M$ emergency vehicles are generated at random time steps, each following a random predefined route through the road network.
Each intersection is controlled by a distributed LLM agent, which dynamically selects appropriate signal phases based on traffic conditions. 
The objective is to optimize signal phase decisions across all intersections to allow emergency vehicles to pass through as quickly as possible and improve the overall traffic condition.



\section{Emergency-Aware Reasoning Framework} \label{sectionIII}

\subsection{Reviewer-Based Emergency RAG}

To enhance adaptive emergency-response decision-making, we design RERAG, an advanced module that integrates an LLM-based reviewer, as shown in Fig. \ref{fig1} (c). 
Equipped with the reviewer, RERAG distils critical knowledge and guidance from historical emergency scenarios and expert-driven emergency response, enabling TSC agents to generate rational and reliable strategies under current traffic conditions.
The proposed RERAG is composed of the following four components. 

\subsubsection{LLM-Based Reviewer} The reviewer $Rev$ is responsible for extracting knowledge from a raw case base $\mathcal{B}$ of historical emergency scenarios to form a structured and generalizable guidance repository. This paper employs a GPT-4o Mini as the reviewer to summarize expert-driven responses. Specifically, the input to $Re$ is a historical case base including intersection $i$ traffic representation of current and next time steps, emergency vehicle states, expert-driven signal phase selection, $b_l=[obs^i_t,Ev_t,a^i_t,obs^i_{t+1},Ev_{t+1}]$. Guided by well-designed prompts, the reviewer $Rev$ identifies the conditions, recommended actions, and intended effects of the guidance by analyzing control strategies and causal relationships in the case base.
The reviewer outputs a set of guidance items $g_d$. The process is formally expressed as $\mathcal{G}=Re(\{b_1,...,b_L\})$, where $L$ and $D$ are the lengths of the historical emergency case base and the guidance repository, respectively.

\subsubsection{Embedding Module} 

The embedding module bridges the guidance repository and retrieval by converting reviewer-generated guidance $g_d$ into a vector, $v_{g_d} = Em(g_d)$.
This paper adopts a sparse mixture-of-experts-based complex text encoder with dynamic position encoding as the embedding module. 
It is able to understand the relative positioning of information in extended sequences.

\subsubsection{LLM-Based Query Generator}
The query generator $QGen$, built upon GPT-4o Mini, interprets the current traffic context and generates semantically rich queries for knowledge retrieval. Given the intersection state $obs^i_{t}$ and emergency vehicle state $Ev_{t}$, $QGen$ formulates a concise query $q_t$ that highlights key features of the emergency situation. 
The generation process is defined as $q_t = QGen(obs^i_{t}, Ev_{t})$.


\subsubsection{Retrieval Module}
The retrieval module identifies the top-$K$ most relevant guidance items from the vectorized guidance repository $v_{\mathcal{G}}=\{v_{g_1},...,v_{g_D}\}$ based on the current query $q_t$.
The query is embedded as $v_{q_t} = Em(q_t)$ using the same encoder as for guidance items. The similarity between $v_q$ and $v_{g_d}$ is computed via cosine similarity,
\begin{flalign}
sim(v_{q_t}, v_{g_d}) = \frac{v_q \cdot v_{g_d}}{|v_q| |v_{g_d}|}.
\label{eq_1}
\end{flalign}
Based on similarity scores, the relevant guidance items $\mathcal{RG}_{t}^i = \{g_{d_1},...,g_{d_K}\}$ are retrieved to assist the signal phase decision. The retrieval process can be formalized as $\mathcal{RG}_{t}^i= Retrieve(v_q, v_{\mathcal{G}})$.
This ensures that the decision is informed by emergency guidance aligned with the current scenario.


\subsection{RERAG-Based Emergency Response with Deep Reasoning}

To handle complex and time-critical emergency scenarios, the emergency-aware reasoning framework enables REG-TSC to perform deep reasoning with the guidance retrieved by RERAG.
The deep reasoning enhances the interpretability and reliability of emergency response, via comprehensively analyzing the current intersection state and retrieved guidance.

The deep reasoning process is selectively triggered under emergency-aware conditions. Specifically, it is enabled only when an emergency vehicle is either currently at intersection $i$, or when intersection $i$ appears in the planned path of the emergency vehicle. This selective activation ensures that deep reasoning is used where it is required most, while reducing unnecessary computational overhead during regular scenarios.

For emergency scenarios, the emergency-aware reasoning framework firstly leverages RERAG to retrieve relevant guidance $\mathcal{RG}_{t}^i$. The retrieved guidance, together with the current traffic representation $obs^i_t$ and emergency vehicle state $Ev_t$, is then fed into the LLM-based signal optimization agent to conduct deep reasoning. The emergency vehicle state $Ev_t$ includes its planned route, its current position, and its current speed. 

Within the LLM-based signal optimization agent, a carefully constructed prompt guides the deep reasoning, as shown in Fig. \ref{fig1} (d). The deep reasoning follows a three-step CoT, \textbf{\textit{Step 1:}} analyzing the current intersection state and the information of the emergency vehicle with guidance, \textbf{\textit{Step 2:}} evaluating each traffic signal phase by predicting emergency vehicle arrival time and future queue lengths, \textbf{\textit{Step 3:}} determining the appropriate phase selection and providing explanation. The deep reasoning process is represented as
\begin{equation}
Ana_t^i,Pre_t^i,A^i_t = \pi^{Emer}_\theta(obs^i_t, Ev_t, \mathcal{RG}_{t}^i),
\label{eq_4}
\end{equation}
where $Ana_t^i$, $Pre_t^i$ and $A^i_t$ denote analysis, prediction and phase decision with explanation.
Deep CoT focuses REG-TSC on the emergency vehicle, enabling fine-grained reasoning via analysis and prediction. 
To reduce overhead, REG-TSC performs lightweight reasoning in regular scenarios, generating a decision with explanation, $A^i_t = \pi^{Regu}_\theta(obs^i_t)$.


\section{Distributed LLM Agents in Generalizable TSC}
\label{sectionIV}

\subsection{Generalizable TSC for Heterogeneous Intersections}

To enable generalization across diverse intersections, we design a type-agnostic traffic representation encoding both static and dynamic features.
The representation $obs^i_t$ is constructed from a lane-centric perspective, where each intersection is decomposed into lanes, allowing every agent to reason consistently across structural and action-space heterogeneity. 
Specifically, $obs^i_t$ is composed of the following four parts.
\begin{itemize}
\item  \textbf{Intersection topology} describes the shape of intersection $i$, and the traffic movements defined among lanes. 
\item  \textbf{Action space} is specified as a set of signal phases, each mapped to its controlled traffic movements.
\item  \textbf{Queuing vehicles} refer to the number of vehicles stopped or slowly moving on lanes controlled by each phase.
\item  \textbf{Approaching vehicles} denote vehicles moving faster than a threshold $v_{stop}$ within a certain segment $s$ of each lane.
\end{itemize}

 
In addition to $obs^i_t$, the prompt fed to the LLMs includes a concise task description and commonsense knowledge to guide the generation of optimal control strategies. A detailed example of the prompt is provided in Appendix \ref{PG}.


The LLM-based signal optimization agent outputs the selected signal phase along with a reasoning trajectory. The phase decision is explicitly marked in the output, allowing REG-TSC to easily extract the action $a^i_t$ and forward it to the traffic simulator. The simulator then returns a reward $r^i_t$ and transitions to the next state $obs^i_{t+1}$, enabling ongoing interaction and learning. The reward $r^i_t$ is defined as,
\begin{equation}
\mathop r\nolimits_t^i  = \mathop \lambda \nolimits_1 \frac{{\mathop {QL}\nolimits_t^i  - \mathop {QL}\nolimits_{t + 1}^i }}{{\mathop {QL}\nolimits_{t + 1}^i }} + \mathop \lambda \nolimits_2 \frac{{\tau  -  {WTE}_t^i }}{{{WTE}_t^i  + \gamma }},
\end{equation}
where ${QL}_t^i$ and ${WTE}_t^i$ denote the queue length at intersection $i$ and the waiting time of emergency vehicle during phase $a^i_t$.

\subsection{Simulation-Driven Fine-Tuning}
\subsubsection{Imitation Fine-Tuning}
This paper performs imitation fine-tuning on Llama-3.1-8B to specialize a general LLM into a TSC agent by imitating high-quality decisions.

Firstly, GPT-4o Mini interacts with a traffic simulator to collect reasoning trajectories. For intersection $i$, GPT-4o Mini generates a reasoning trajectory $\widehat{Y^{i}_t}$ with a CoT and selected action according to prompt $X^{i}_t$. To ensure data quality, we apply a \textit{reward-based filter} that retains only effective trajectories aligned with long-term traffic optimization. Specifically, a trajectory is preserved if the cumulative reward over the next $t_{re}$ steps exceeds a predefined threshold $\eta$, i.e. $\sum_{k=0}^{t_{re}-1} r^i_{t+k} \geq \eta$.
These high-quality trajectories are stored into a fine-tuning dataset.
Subsequently, we conduct off-line supervised training by minimizing the negative log-likelihood (NLL) loss,
\begin{equation}
\mathcal{L}_{\text{NLL}} = -\sum_{\omega=1}^{|\widehat{Y^{i}_t}|} \log P_{\pi_\theta}(\widehat{Y^{i}_{t,\omega}} \mid X^{i}_t, \widehat{Y^{i}_{t,{<\omega}}}),
\label{eq_5}
\end{equation}
where $\widehat{Y^{i}_{t,\omega}}$ is the $\omega$-th token in the response. Low-rank adaptation is adopted for parameter-efficient fine-tuning. The process transfers GPT’s reasoning logic to REG-TSC, providing a TSC-specialized initialization for subsequent training.

\subsubsection{Reward-Guided Reinforced Refinement}


To boost adaptability to diverse intersections, we propose $\text{R}^3$, which adaptively focuses training on challenging intersection types.

In $\text{R}^3$, REG-TSC interacts with the simulator across $N$ heterogeneous intersection types. At time step $t$, REG-TSC generates a reasoning trajectory $Y^i_t$ according to a prompt $X^i_t$ and receives a reward $r^i_t$ from the simulator. The tuple $(X^i_t, Y^i_t, r^i_t)$ is then stored into the experience buffer $\mathcal{B}_n$ corresponding to intersection type $n$. 
To emphasize low-performing types, we define the sampling probability $SPr_n$ as
\begin{equation}
SP{r_n} = \frac{{1/({{\bar r}_n} + \left| {\min (\bar r)} \right| + \varepsilon )}}{{\sum\limits_{k = 1}^N {(1/({{\bar r_k}} + \left| {\min (\bar r)} \right| + \varepsilon ))} }}.
\label{eq_6}
\end{equation}
Prioritized samples drawn from all experience buffers are combined into a refinement dataset. A reward-weighted NLL loss is employed to guide the refinement,
\begin{equation}
\mathcal{L}_{\text{RNLL}} = -\sum_{(X^i_t, Y^i_t, r^i_t)} r^i_t \cdot \sum_{\omega=1}^{|Y^i_t|} \log P_{\pi_\theta}(Y^i_{t,\omega} \mid X, Y^i_{t,<\omega}).
\label{eq_7}
\end{equation}



\subsection{REG-TSC Learning and Reasoning Process}

The learning and reasoning process of REG-TSC is shown in Algorithm \ref{alg_1}. 
Firstly, REG-TSC is initialized and trained with trajectories generated by GPT interacting with a traffic simulator. These trajectories are filtered via \textit{reward-based filter} to build a high-quality imitation dataset. Secondly, during reinforced refinement, emergency intersections retrieve guidance and perform deep reasoning accordingly, while intersections in regular scenarios follow the lightweight reasoning. After executing actions and observing rewards, the transitions are stored in corresponding experience buffers. REG-TSC is progressively refined with prioritized samples.

\begin{algorithm}[h]\label{alg_1}
            \caption{REG-TSC Learning and Reasoning}          
            \LinesNumbered
            
            Initialize REG-TSC and imitation dataset $\cal D$ \;

            Interact GPT with a traffic simulation environment \; 
            Collect reasoning trajectory $(X^{i}_t,\widehat{Y^{i}_t})$ and store to $\cal D$\;
            Filter $\cal D$ with reward-based filter\;
            Fine-tune REG-TSC with $\cal D$ and Eq. \eqref{eq_5}\;
          
           Initialize experience buffers $\{\mathcal{B}_n\}$ and vectorize $\mathcal{G}$\; 

           \For{e=1:max\_epoch}{
           Initialize a traffic simulation environment\; 
           \For{t=1:T}{
           
           \For{i=1:I}{
           \If{ $i$ in emergency scenario}{
            Get $obs^i_t, Ev_t$ from the environment\;
            Get query and retrieve $\mathcal{RG}_{t}^i$ via Eq. \eqref{eq_1}\;
            Reason $Ana_t^i,Pre_t^i,A^i_t$ via Eq. \eqref{eq_4}\;      
           }
           \Else{
           Get $obs^i_t$ from the environment\;
           Reason $A^i_t$ via $\pi^{Regu}_\theta(obs^i_t)$\;
           }
           }
           Extract $\{a_t^i\}$ to perform and observe $\{r_t^i\}$\;
           Append $(X^i_t, Y^i_t, r^i_t)$ to corresponding $\mathcal{B}_n$\;
           }
           Sample training data from $\{\mathcal{B}_n\}$ via Eq. \eqref{eq_6}\;
           Perform reinforced refinement via Eq. \eqref{eq_7}\;
          
           }
\end{algorithm}

\section{Experiments and Results}
\label{sectionV}
\subsection{The Simulator and Settings}
\begin{figure}[H]
\centering
\includegraphics[width=0.95\columnwidth]{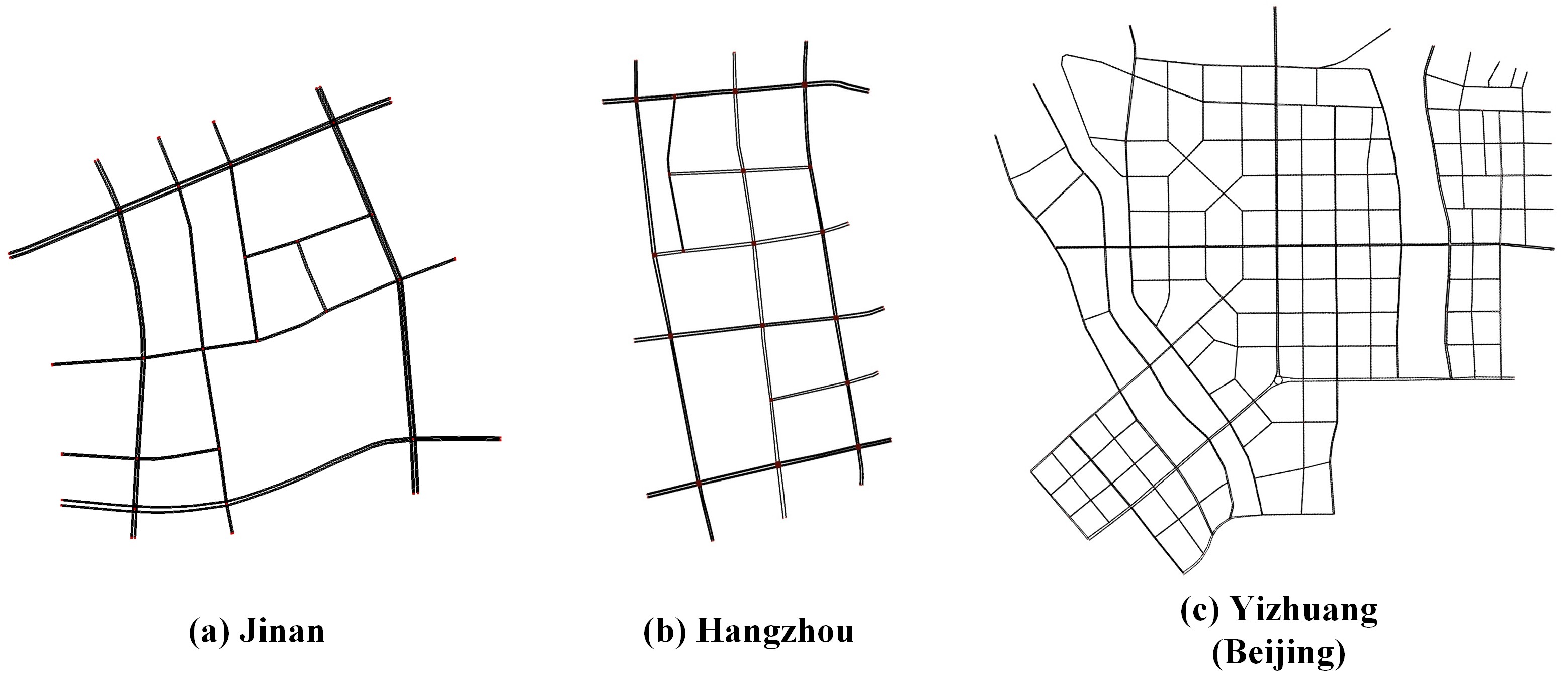}
\caption{Simulated Urban Road Networks.}
\label{network}
\end{figure}

To thoroughly validate the effectiveness and robustness of REG-TSC, we construct three road networks based on SUMO \cite{EX1}, corresponding to parts of the roads in Jinan, Hangzhou, and Yizhuang (Beijing), as shown in Fig. \ref{network}. The parameters of traffic flow datasets are presented in TABLE \ref{Datasets}. The settings of REG-TSC and simulator are shown in TABLE \ref{settings}. The evaluation of methods is based on the following metrics: the average travel time (ATT) and average waiting time (AWT) of all vehicles, the average queue length (AQL) at all intersections, and the average travel time (ATTE) and average waiting time (AWTE) of emergency vehicles.
\begin{table}[H]
\caption{Traffic Flow Datasets}
\label{Datasets}
\begin{center}
\begin{tabular}{c|c|c|c}
\toprule
Traffic Flow Dataset     & \begin{tabular}[c]{@{}c@{}}Intersection\\ Number $I$\end{tabular} & \begin{tabular}[c]{@{}c@{}}Vehicle\\ Number\end{tabular} & \begin{tabular}[c]{@{}c@{}}Arrival Rate\\ (vehicle/min)\end{tabular} \\ \hline
Jinan1           & \multirow{2}{*}{17}                                           & 1710                                                     & 57.14                                                                \\
Jinan2           &                                                               & 2400                                                     & 80.00                                                                \\ \hline
Hangzhou1        & \multirow{2}{*}{19}                                           & 1740                                                     & 58.03                                                                \\
Hangzhou2        &                                                               & 2440                                                     & 81.30                                                                \\ \hline
Yizhuang1        & \multirow{3}{*}{177}                                          & 3600                                                     & 120.00                                                               \\
Yizhuang2        &                                                               & 8000                                                     & 266.67                                                               \\
Yizhuang Extreme &                                                               & 10500                                                    & 350.88                                                               \\ \bottomrule
\end{tabular}
\end{center}
\end{table}
\begin{table}[H]
\caption{Algorithm and Simulation Parameters}
\centering
\begin{tabular}{cc|cc}
\toprule
\textbf{Parameters} & \textbf{Value} & \textbf{Parameters} & \textbf{Values} \\ \hline
       $\lambda_1$ &    5   & $\lambda_2$&   1\\

       $\tau$ &      5    &  $\gamma$ &      1       \\ 
       $\eta$ &  0.5    & $\varepsilon$ &  0.1     \\
        $K$ &   1    &   Learning Rate &  $3\times10^{-4}$       \\
       $T$& 1800 & $M$ &   6 \\ 
\bottomrule 
\end{tabular}
\label{settings}
\end{table}

\subsection{Ablation Experiments}
Ablation experiments are conducted to explore the effects of emergency-aware reasoning framework and $\text{R}^3$, shown in the last three columns of TABLE \ref{tabel_result}. Method A represents REG-TSC only trained via imitation fine-tuning without $\text{R}^3$. Method B represents REG-TSC without emergency-aware reasoning. 

REG-TSC outperforms Method A under varying traffic conditions. In Jinan2, REG-TSC reduces ATT by 17.92\% and AQL by 14.91\% compared with Method A. In Yizhuang network, the AWT and AWTE of Method A are 21.52\% and 38.96\% more than those of REG-TSC on average. Therefore, $\text{R}^3$ serves an essential role in optimizing policies across heterogeneous intersections.

Method B shows a limited effectiveness in handling emergency scenarios. Across the tested six scenes, AWTE of Method B exceeds that of REG-TSC by 86.13\% on average. Specifically, in Hangzhou1, Method B records an AWTE 82.00 seconds longer than REG-TSC. It indicates that the emergency-aware reasoning framework enables REG-TSC to deliver reliable decisions in emergency scenarios.

\begin{figure}[H]
\centering
\includegraphics[width=\columnwidth]{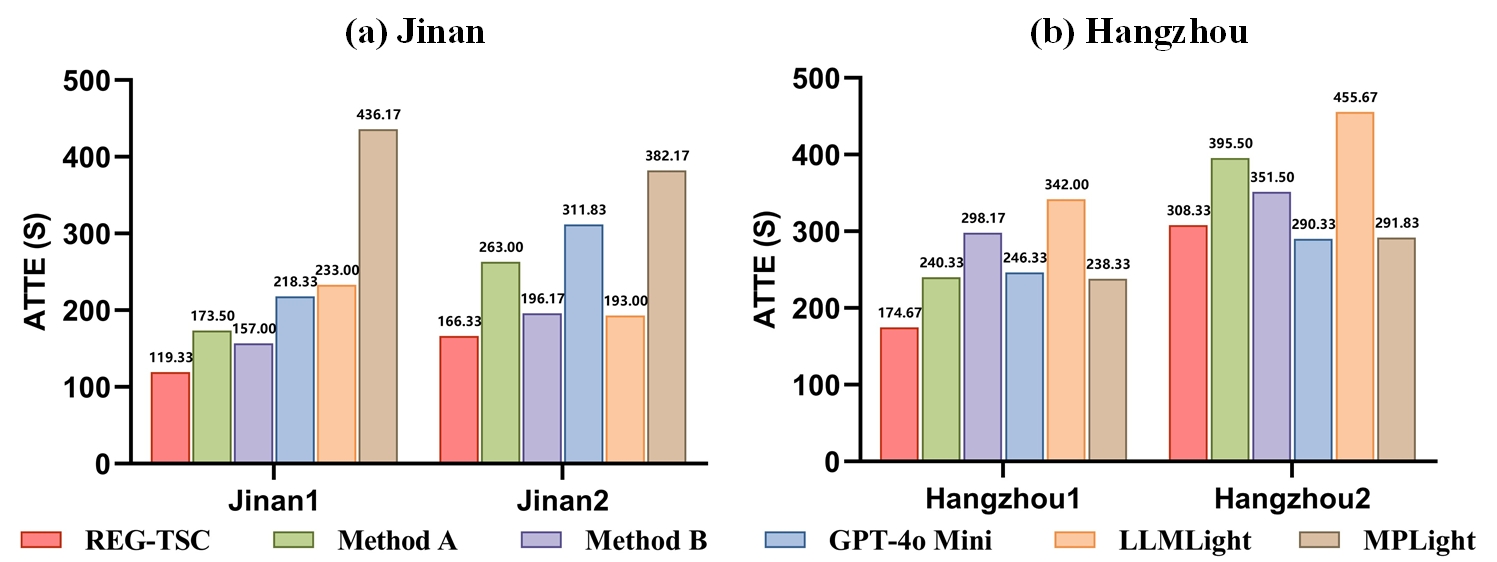}
\caption{ATTE on Jinan and Hangzhou Road Networks.}
\label{atte}
\end{figure}
Fig. \ref{atte} provides a clear comparison of ATTE. In Jinan1, REG-TSC shortens ATTE by 54.17 seconds compared with Method A.
Equipped with emergency-aware reasoning, REG-TSC achieves 12.28\% shorter ATTE than Method B in Hangzhou2.
Across all scenes, REG-TSC maintains the shortest ATTE, demonstrating strong emergency response ability.

In addition, in Jinan1, the average inference time of REG-TSC is 4.07 seconds per time step, similar to that of GPT-4o Mini measured at 3.85 seconds. This shows that REG-TSC has excellent deployability and low computational overhead.

\begin{table*}[]
\caption{Evaluation Results}
\label{tabel_result}
\begin{center}
\resizebox{\textwidth}{!}{
\begin{tabular}{cc|cccccc|cccc|ccc}
\toprule
\multicolumn{2}{c|}{Methods}      & MPLight & AttendLight & PressLight & CoLight & \begin{tabular}[c]{@{}c@{}}Efficient-\\ Colight\end{tabular} & \begin{tabular}[c]{@{}c@{}}Advanced-\\ Colight\end{tabular} & \begin{tabular}[c]{@{}c@{}}DeepSeek-\\ R1-8B\end{tabular} & \begin{tabular}[c]{@{}c@{}}GPT-\\ 4o Mini\end{tabular} & LLMLight        & \begin{tabular}[c]{@{}c@{}}Llama 3.1\\ -8B\end{tabular} & \begin{tabular}[c]{@{}c@{}}Method\\ A\end{tabular} & \begin{tabular}[c]{@{}c@{}}Method\\ B\end{tabular} & REG-TSC         \\ \hline
\multirow{4}{*}{Jinan1}    & ATT  & 272.63  & 811.42      & 275.19     & 795.17  & 709.13                                                       & 677.13                                                      & 309.63                                                    & 245.17                                                 & 222.10          & 274.64                                                  & 266.38                                             & 231.74                                             & \textbf{214.04} \\
                           & AWT  & 165.03  & 773.99      & 168.79     & 746.43  & 590.99                                                       & 611.58                                                      & 202.82                                                    & 132.24                                                 & \textbf{106.23} & 167.48                                                  & 158.22                                             & 120.13                                             & 110.10          \\
                           & AQL  & 20.72   & 72.40       & 21.10      & 54.18   & 19.37                                                        & 52.30                                                       & 19.76                                                     & 12.84                                                  & 11.71           & 16.84                                                   & 15.23                                              & 12.30                                              & \textbf{11.56}  \\
                           & AWTE & 320.50  & 1081.67     & 132.33     & 1039.17 & 534.17                                                       & 1121.83                                                     & 128.17                                                    & 103.33                                                 & 111.50          & 369.67                                                  & 62.50                                              & 29.50                                              & \textbf{11.83}  \\ \hline
\multirow{4}{*}{Jinan2}    & ATT  & 301.82  & 743.30      & 289.66     & 797.97  & 768.17                                                       & 716.03                                                      & 378.27                                                    & 296.97                                                 & 248.35          & 360.00                                                  & 290.70                                             & 254.46                                             & \textbf{238.62} \\
                           & AWT  & 198.13  & 689.88      & 186.79     & 740.38  & 699.52                                                       & 659.28                                                      & 256.60                                                    & 176.13                                                 & 130.94          & 244.67                                                  & 173.46                                             & 142.00                                             & \textbf{127.90} \\
                           & AQL  & 24.72   & 73.66       & 23.60      & 57.35   & 56.85                                                        & 59.99                                                       & 34.57                                                     & 22.04                                                  & \textbf{18.34}  & 29.31                                                   & 21.66                                              & 20.46                                              & 18.43           \\
                           & AWTE & 300.83  & 609.83      & 154.67     & 755.17  & 1054.83                                                      & 905.17                                                      & 175.50                                                    & 172.50                                                 & 81.33           & 442.00                                                  & 133.33                                             & 63.17                                              & \textbf{54.50}  \\ \hline
\multirow{4}{*}{Hangzhou1} & ATT  & 262.76  & 800.02      & 278.23     & 762.18  & 526.24                                                       & 611.79                                                      & 382.95                                                    & 276.75                                                 & 258.98          & 315.97                                                  & 270.04                                             & 254.29                                             & \textbf{246.72} \\
                           & AWT  & 136.53  & 740.82      & 152.14     & 701.25  & 422.16                                                       & 523.49                                                      & 243.86                                                    & 142.01                                                 & 124.17          & 189.07                                                  & 135.55                                             & 121.48                                             & \textbf{120.28} \\
                           & AQL  & 16.17   & 57.57       & 18.66      & 57.52   & 45.65                                                        & 46.58                                                       & 22.33                                                     & 12.94                                                  & 12.38           & 17.39                                                   & 12.31                                              & 11.69                                              & \textbf{11.13}  \\
                           & AWTE & 112.00  & 1008.33     & 145.50     & 1245.50 & 269.17                                                       & 593.33                                                      & 154.83                                                    & 81.33                                                  & 184.33          & 287.33                                                  & 93.00                                              & 110.83                                             & \textbf{28.83}  \\ \hline
\multirow{4}{*}{Hangzhou2} & ATT  & 312.55  & 794.47      & 297.85     & 839.87  & 522.19                                                       & 655.17                                                      & 387.85                                                    & 292.91                                                 & 290.29          & 349.89                                                  & 339.19                                             & 303.50                                             & \textbf{284.25} \\
                           & AWT  & 176.27  & 740.62      & 165.14     & 792.94  & 411.64                                                       & 560.66                                                      & 252.80                                                    & 161.20                                                 & \textbf{148.74} & 215.69                                                  & 205.32                                             & 175.82                                             & 152.57          \\
                           & AQL  & 25.81   & 93.01       & 32.12      & 81.72   & 51.63                                                        & 67.93                                                       & 28.56                                                     & 20.98                                                  & \textbf{18.67}  & 26.44                                                   & 24.56                                              & 20.77                                              & 20.30           \\
                           & AWTE & 129.17  & 1338.33     & 240.50     & 1269.67 & 534.17                                                       & 651.00                                                      & 128.50                                                    & \textbf{111.17}                                        & 246.67          & 168.00                                                  & 225.83                                             & 149.83                                             & 129.67          \\ \hline
\multirow{4}{*}{Yizhuang1} & ATT  & 462.94  & 859.63      & 497.51     & 868.73  & 436.01                                                       & 830.12                                                      & 597.92                                                    & 470.16                                                 & 452.59          & 485.87                                                  & 454.42                                             & 427.45                                             & \textbf{401.59} \\
                           & AWT  & 231.88  & 796.51      & 274.59     & 792.75  & 343.45                                                       & 721.02                                                      & 378.94                                                    & 212.82                                                 & 193.43          & 259.40                                                  & 196.62                                             & 177.11                                             & \textbf{169.80} \\
                           & AQL  & 13.29   & 26.16       & 14.27      & 22.62   & 40.30                                                        & 24.44                                                       & 6.17                                                      & 3.82                                                   & 3.57            & 4.76                                                    & 3.55                                               & 3.46                                               & \textbf{3.43}   \\
                           & AWTE & 186.67  & 804.17      & 118.00     & 305.67  & 373.17                                                       & 644.17                                                      & 351.50                                                    & 218.33                                                 & 168.67          & 452.83                                                  & 252.33                                             & 146.24                                             & \textbf{127.45} \\ \hline
\multirow{4}{*}{Yizhuang2} & ATT  & 545.12  & 882.65      & 662.68     & 872.58  & 754.29                                                       & 863.98                                                      & 561.82                                                    & 561.75                                                 & 521.01          & 580.25                                                  & 537.46                                             & 517.71                                             & \textbf{478.30} \\
                           & AWT  & 327.17  & 830.13      & 469.63     & 803.09  & 578.01                                                       & 790.60                                                      & 341.41                                                    & 292.53                                                 & 256.42          & 337.13                                                  & 318.73                                             & 291.86                                             & \textbf{225.00} \\
                           & AQL  & 27.61   & 43.77       & 31.11      & 41.74   & 33.36                                                        & 42.14                                                       & 14.17                                                     & 11.11                                                  & \textbf{10.24}  & 12.39                                                   & 10.56                                              & 10.47                                              & 10.32           \\
                           & AWTE & 224.17  & 1310.00     & 665.00     & 1269.67 & 732.83                                                       & 1260.67                                                     & 710.83                                                    & 264.67                                                 & 373.83          & 615.17                                                  & 184.67                                             & 180.83                                             & \textbf{132.17} \\ \bottomrule
\end{tabular}
} 
\end{center}
\end{table*}

\subsection{Comparison with Other TSC Methods}

We compare REG-TSC with six RL-based approaches: MPLight, AttendLight \cite{HI2}, PressLight, CoLight, Efficient-CoLight, and Advanced-CoLight; and a SOTA LLM-based method: LLMLight \cite{RAG10}. Furthermore, we evaluate the performance of general LLMs integrated within our emergency-aware reasoning framework, including Deepseek-R1-8B, GPT-4o Mini, Llama 3.1-8B. The results are shown in TABLE \ref{tabel_result}.

REG-TSC consistently achieves SOTA or comparable performance against all baselines, highlighting outstanding emergency vehicle handling and strong generalization capabilities. Specifically, in Hangzhou1, REG-TSC achieves 3.13\% and 10.10\% lower AWT and AQL, respectively, compared with LLMLight, the second-best method. For the complex Yizhuang2, REG-TSC reduces AWTE by 69.99\% on average, surpassing the other three general LLMs. Moreover, LLM-based methods basically outperform RL-based methods, indicating LLM's superior generalization in optimizing signal phases across heterogeneous intersections. Notably, by learning from specific scenes and traffic data, MPLight and PressLight can effectively develop pressure-based signal phase policies, achieving reasonable decision-making. However, their performance still lags behind REG-TSC. For example, across six scenes, MPLight’s ATT and AWTE are on average 13.85\% and 54.14\% higher than those of REG-TSC.

The results in Fig. \ref{atte} highlight REG-TSC’s superior ability to facilitate rapid emergency vehicle passage compared with both LLM-based and traditional baselines.
For example,in Hangzhou1, REG-TSC reaches an ATTE of 174.67 seconds, outperforming GPT-4o Mini and MPLight by margins of 71.66 seconds and 63.66 seconds, respectively.
\begin{figure}[H]
    \centering
      \includegraphics[width=0.60\columnwidth]{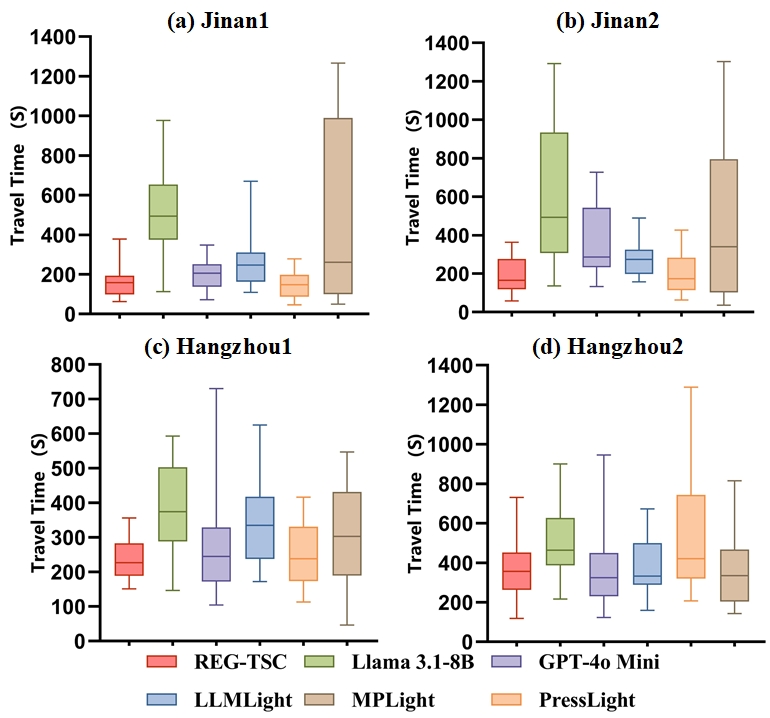}
    \caption{Travel Time of Emergency Vehicles ($M=20$).}
    \label{tte}
\end{figure}
To further evaluate the ability of different methods to handle emergency scenarios, we set the number of emergency vehicles to 20 and measure their arrival time on the Jinan and Hangzhou road networks, as shown in Fig. \ref{tte}. In Jinan1, REG-TSC achieves a median travel time of around 180 seconds, which is substantially shorter than Llama 3.1-8B at about 450 seconds and GPT-4o Mini at about 260 seconds. Moreover, REG-TSC consistently exhibits small variances in travel time distributions across all scenes. These results show that REG-TSC outperforms other methods in the rationality and reliability of decision-making when handling emergency vehicles.

\subsection{Generalization Comparison}

\begin{figure}[h]
\centering
\includegraphics[width=0.95\columnwidth]{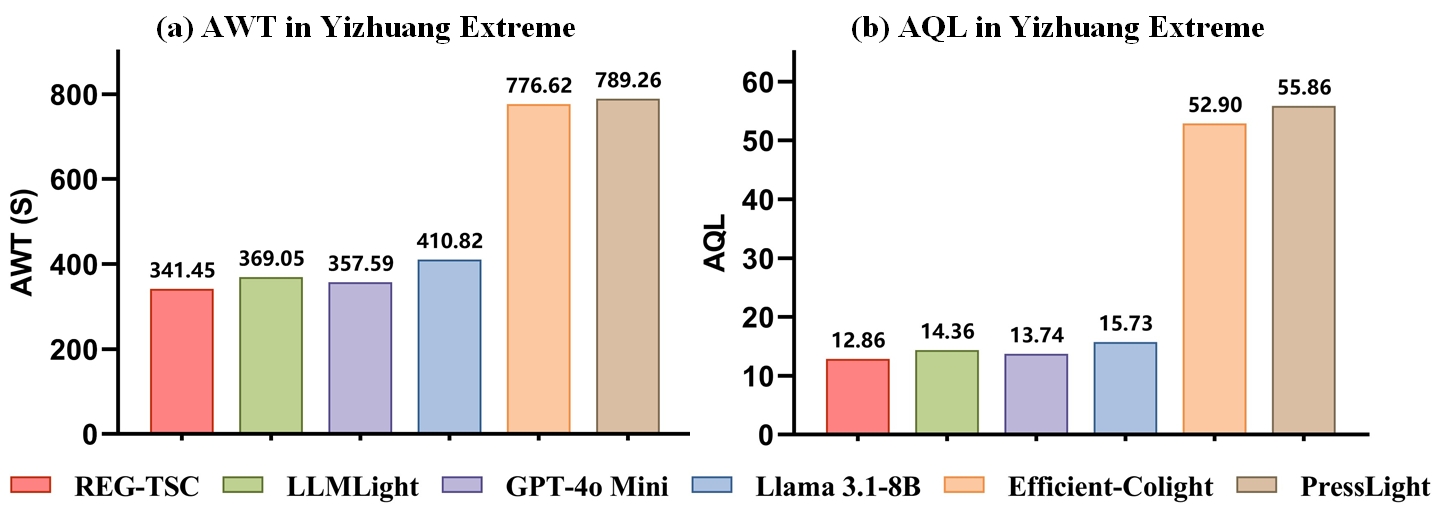}
\caption{Performance Under Extreme Traffic Conditions ($M=0$). }
\label{extrem}
\end{figure}

To evaluate the generalization capability of different methods, we train all models on the Hangzhou2 dataset except general LLMs and test them on the Yizhuang Extreme, as shown in Fig. \ref{extrem}. Compared with RL-based methods, the LLM-based methods show stronger generalization, making reasonable decisions even in unseen environments. REG-TSC achieves the lowest AWT of 341.45 seconds and the shortest AQL of 12.86, significantly outperforming all baselines. In contrast, RL-based methods exhibit much higher AWT above 775 seconds and longer AQL exceeding 50. The results confirm that REG-TSC is able to generalize well to unfamiliar and extreme traffic conditions with outstanding performance.

\section{Conclusion}
\label{sectionVI}
TSC plays a pivotal role in optimizing urban traffic. This paper proposes REG-TSC for generalizable signal optimization with emergency response. The emergency-aware reasoning framework enables REG-TSC to make reasonable and reliable decisions in emergency scenarios. $\text{R}^3$  adaptively trains REG-TSC across heterogeneous intersections, enhancing generalization capability. Extensive experiments in 6 traffic scenes show that REG-TSC reduces ATT by 42.00\% and AWTE by 83.16\% compared with other SOTA methods. In the future, we will consider mixed autonomous driving scenarios and study vehicle–infrastructure cooperation for LLM-based TSC.


\bibliographystyle{IEEEtran}
\bibliography{conference_101719}

\onecolumn
\appendices

\section{Prompt Template and Response Sample of REG-TSC}  
\subsection{Prompt Template for REG-TSC} \label{PG}

\begin{table}[H]
\centering
\caption{Prompt Template for REG-TSC}
\begin{tabular}{p{\textwidth}}
\toprule
\textbf{[Role]} \\
You are a traffic signal control agent with emergency response. \\
\hline
\textbf{[Objective]} \\
Based on the real-time traffic representation, emergency vehicle state, critical guidance for emergency scenarios and commonsense knowledge, determine the most effective next traffic signal phase that allows emergency vehicles to pass through as quickly as possible and improve the overall traffic condition. \\
\hline
\textbf{[Real-Time Traffic Representation and Emergency Vehicle State]} \\
\textbf{Intersection Topology}: There are 4 bidirectional roads connected to this intersection (ID: road\#1, road\#2, road\#3, road\#4), with 2, 2, 2, 2 incoming lanes respectively. 
A total of 16 traffic movements are managed by 4 signal phases in this intersection.
\\
\textbf{Action Space:} (Traffic movements allowed by each phase, upstream lane→downstream lane): \\
Phase 1: road\#1\_2 → -road\#3\_2; road\#1\_2 → -road\#4\_2; road\#3\_2 → -road\#1\_2; road\#3\_2 → -road\#2\_2  \\
Phase 2: road\#1\_1 → -road\#2\_1; road\#1\_1 → -road\#1\_1; road\#3\_1 → -road\#4\_1; road\#3\_1 → -road\#3\_1 \\
Phase 3: road\#2\_2 → -road\#4\_2; road\#2\_2 → -road\#1\_2; road\#4\_2 → -road\#2\_2; road\#4\_2 → -road\#3\_2 \\
Phase 4: road\#2\_1 → -road\#3\_1; road\#2\_1 → -road\#2\_1; road\#4\_1 → -road\#1\_1; road\#4\_1 → -road\#4\_1 \\
\textbf{Queuing and Approaching Vehicles}:
The number of queuing vehicles (QV) is counted on upstream lanes controlled by each signal phase. Each lane is divided into three equal-length segments from the lane start to the stop line, representing the far, middle, and near sections to the stop line. The count of approaching vehicles (AV) in each segment is recorded accordingly as far/mid/near. \\
\begin{tabular}{p{0.23\textwidth} p{0.23\textwidth} p{0.23\textwidth} p{0.23\textwidth}}
\textbf{Phase 1} & \textbf{Phase 2} & \textbf{Phase 3} & \textbf{Phase 4} \\
road\#1\_2: QV=4; AV=0/3/0 & road\#1\_1: QV=2; AV=1/1/0 & road\#2\_2: QV=3; AV=1/0/2 & road\#2\_1: QV=5; AV=2/1/1 \\
road\#3\_2: QV=1; AV=3/0/2 & road\#3\_1: QV=1; AV=2/0/1 & road\#4\_2: QV=2; AV=1/1/3 & road\#4\_1: QV=2; AV=1/2/0 \\
Total: QV=5; AV=3/3/2 & Total: QV=3; AV=3/1/1 & Total: QV=5; AV=2/1/5 & Total: QV=7; AV=3/3/1 
\end{tabular} \\
\textbf{Emergency Vehicle State}: \\
Emergency Vehicle ID: Ambulance\_1 \\
Planned Route: road\#72→ road\#64→ road\#2→ road\#4→road\#9→road\#32 \\
Current Position: road\#2\_1, $276.8m$ to stop line \\
Speed: $17.4m/s$ \\
\hline
\textbf{[Critical Guidance for Emergency Scenarios]} \\
Current Possible Situation: An emergency vehicle is approaching the intersection, but its lane is still occupied by queuing vehicles. \\
Recommended Action: Promptly select the signal phase for the lane with the emergency vehicle.\\
Intended Effect: Clear the queuing vehicles in the lane with the emergency vehicle for it rapid passage.\\
\hline
\textbf{[Commonsense Knowledge]} \\
1. EMERGENCY VEHICLE PRIORITY: Emergency vehicles have the highest priority. Phase selection should PRIMARILY aim to minimize their waiting time and allow them to pass through the intersection as quickly as possible. \\
2. MAXIMIZE THROUGHPUT: Choose phases that reduce overall traffic delay and congestion across all lanes. \\
3. EARLY QUEUE URGENCY: Traffic congestion at intersections is mostly caused by vehicles queued NEAR the stop line. PRIORITIZE lanes with long queues there, while vehicles in distant segments can wait. \\
4. DOWNSTREAM BLOCKAGE CAUTION: Avoid activating any lane that would push the downstream link near capacity, as this risks spill back and network-wide delays. \\
5. WAITING TIME FAIRNESS: Lanes that have waited excessively must be served once downstream allows—they cannot be skipped indefinitely. \\
6. LANE RULES: Vehicles are permitted to pass one at a time per lane. All vehicles, INCLUDING emergency vehicles, must follow the queuing order, meaning a vehicle cannot move until those ahead of it have departed. \\
\hline
\textbf{[Task Description]} Carefully analyze the given information, think step by step, and make well-reasoned decisions.\\
1. \textbf{Analysis}: Interpret and analyze the current traffic conditions and the emergency vehicle state. \\
2. \textbf{Prediction and Evaluation}: Evaluate and compare traffic signal phases by predicting emergency vehicle arrival time at the intersection and future queue lengths. Integrate critical guidance and commonsense knowledge to ensure the emergency vehicle can pass through intersections without delay. \\
3. \textbf{Decision-Making}: Select an appropriate traffic signal phase that enables emergency vehicles to pass through the intersection as quickly as possible and reduces overall traffic congestion. Provide the appropriate phase selection and an explanation. \\
\hline
\textbf{[Output Format]} \\
Your response must strictly follow the format below. \\
\texttt{<response>} \\
\quad \texttt{<traffic analysis>}INSERT\_ANALYSIS\_HERE\texttt{</traffic analysis>} \\
\quad \texttt{<evaluation and explanation>}INSERT\_EVALUATION\_AND\_EXPLANATION\_HERE\texttt{</evaluation and explanation>} \\
\quad \texttt{<signal>}INSERT\_PHASE\_NUMBER\_HERE\texttt{</signal>} \\
\texttt{</response>} \\
\bottomrule
\end{tabular}
\label{tab:tsc_prompt}
\end{table}

\subsection{Response Sample of REG-TSC}
\begin{footnotesize}
\textit{\textless response\textgreater} \textit{\textless traffic analysis\textgreater} 
The emergency vehicle Ambulance\_1 is on road\#2\_1, 276.8 m from the stop line, moving at 17.4 m/s. This lane has 5 queuing vehicles and approaching vehicles distributed as 2/1/1 (far/mid/near). Phase 4 controls this lane, but the lane is currently blocked by queued vehicles. Other phases either do not serve the emergency vehicle's lane or have lower total queue urgency.
\textit{\textless /traffic analysis\textgreater} 
\textit{\textless evaluation and explanation \textgreater}
At current speed, Ambulance\_1 will reach the stop line in 15.9 s. Only Phase 4 will allow clearing the queue ahead. Prioritizing this phase minimizes waiting time for the ambulance. Other phases would delay the emergency vehicle. Phase 4 also has the highest total QV=7. This aligns with emergency vehicle priority, early queue urgency, and overall throughput considerations.
\textit{\textless /evaluation and explanation\textgreater}
\textit{\textless signal\textgreater}4\textit{\textless /signal\textgreater}  \textit{\textless /response\textgreater}
\end{footnotesize}

\end{document}